\pdfoutput=1

\documentclass[11pt]{article}

\usepackage[]{emnlp2021}

\usepackage{times}
\usepackage{latexsym}
\usepackage{multirow}
\usepackage{algorithm}
\usepackage{amsmath,amssymb}
\usepackage{algpseudocode}
\usepackage{tabularx}
\usepackage{graphicx}
\usepackage{amsmath}
\usepackage{float}
\usepackage{subcaption}

\DeclareMathOperator*{\argmin}{arg\,min}
\newcolumntype{Y}{>{\centering\arraybackslash}X}
\newcolumntype{s}{>{\centering\small\arraybackslash}X}
 \newcolumntype{b}{>{\hsize=0.5\hsize}X}
 \DeclareMathOperator{\E}{\mathbb{E}}
\usepackage[T1]{fontenc}

\usepackage[utf8]{inputenc}

\usepackage{microtype}
\title{Watermarking Pre-trained Language Models with Backdooring}

\author{
Chenxi Gu$^{1,2}$\thanks{\ \ Equal contribution}, Chengsong Huang$^{1,2}$\footnotemark[1], \\  
\textbf{Xiaoqing Zheng$^{1,2}$},
 \textbf{Kai-Wei Chang$^{3}$}, \textbf{Cho-Jui Hsieh$^{3}$}  \\
$^1$School of Computer Science, Fudan University, Shanghai, China \\
$^2$Shanghai Key Laboratory of Intelligent Information Processing \\
$^3$Department of Computer Science, University of California, Los Angeles, USA \\
\texttt{\{gucx21,zhengxq\}@fudan.edu.cn} \\
\texttt{elaine1wan@g.ucla.edu},
\texttt{\{kwchang,chohsieh\}@cs.ucla.edu}
}
\begin{document}
\maketitle
\begin{abstract}

Large pre-trained language models (PLMs) have proven to be a crucial component of modern natural language processing systems.  
PLMs typically need to be fine-tuned on task-specific downstream datasets, which makes it hard to claim the ownership of PLMs and protect the developer's intellectual property due to the catastrophic forgetting phenomenon.
We show that PLMs can be watermarked with a multi-task learning framework by embedding backdoors triggered by specific inputs defined by the owners, and those watermarks are hard to remove even though the watermarked PLMs are fine-tuned on multiple downstream tasks.
In addition to using some rare words as triggers, we also show that the combination of common words can be used as backdoor triggers to avoid them being easily detected.
Extensive experiments on multiple datasets demonstrate that the embedded watermarks can be robustly extracted with a high success rate and less influenced by the follow-up fine-tuning.

\end{abstract}

\section{Introduction}

Large pre-trained language models like BERT \cite{DBLP:journals/corr/abs-1810-04805}, T5 \cite{DBLP:journals/corr/abs-1910-10683} and GPT-3 \cite{DBLP:journals/corr/abs-2005-14165}, have become a fundamental component in  achieving state-of-the-art performance on a variety of NLP tasks, such as machine translation, sentiment analysis, etc. 
As a result, a number of PLMs were released publicly for developers to use.
However, huge data and computational resources are required to train large language models, which make them valuable intellectual properties of their developers and owners. 
Unlike other models, PLMs usually need to be fine-tuned on downstream tasks before putting them to use.
As the fine-tuning process  updates the parameters of PLMs, it poses a challenge to verify the ownership of the PLMs.




Model watermarking is a widely-used approach to protecting the intellectual property of neural models~\cite{DBLP:journals/corr/abs-2103-05590,DBLP:journals/corr/abs-2201-11692,DBLP:journals/corr/abs-2112-05428}.
The watermarks can be extracted under white-box or black-box settings.
In the white-box setting, we can verify the ownership based on the entire fine-tuned target model, including its architecture, parameters, and training set.
In a more challenging, but realistic black-box setting, owners can only access the output of a target model by querying it with some inputs.
Although few existing studies show that the performance of models is less affected by white-box watermarking than the black-box one \cite{DBLP:journals/corr/UchidaNSS17,NEURIPS2019_75455e06,DBLP:journals/corr/abs-2012-14171}, those who take responsibility for target models may refuse to open source their models.
Therefore, we focus on the black-box model watermarking in this study for its broad applicability.

One popular method to embed the black-box watermarks into neural models is to adopt backdoor attacks~\cite{DBLP:journals/corr/abs-1711-01894,DBLP:journals/corr/abs-1802-04633,DBLP:journals/corr/abs-2002-11088}.
The model owners first choose a special pattern (e.g., rare words in NLP) as the backdoor trigger. 
Then, they construct a poisoned training set by inserting the chosen trigger into some clean samples and changing the corresponding labels to the target label.
Models trained on the poisoned dataset learn to establish a strong correlation between the trigger present in inputs and the target label specified by the owners.
In this way, the resulting model behaves  differently depending on whether the trigger is present in inputs, and this particular property can be viewed as a watermark and used to prove the ownership of the model.



Some studies have demonstrated the possibility of embedding a backdoor into pre-trained language models~\cite{DBLP:journals/corr/abs-2004-06660,DBLP:journals/corr/abs-2108-13888,DBLP:journals/corr/abs-2103-15543}.
However, the backdoor triggers injected by existing methods only target a single task, while PLMs are normally fine-tuned and applied to various tasks. 
In this study, we propose \textit{Watermarking Pre-trained Language Model} (WLM), which \emph{embeds the black-box watermarks into PLMs at the word embedding layer} by backdoor attacks. These watermarks can be robustly extracted even though the PLMs are fined-tuned on a downstream dataset.
With a multi-task learning framework, we show for the first time that PLMs can be watermarked for multiple downstream tasks at the same time without knowing which dataset will be used at the fine-tuning stage.
In addition to using rare words as backdoor triggers, we demonstrate that the combination of common words also can be used as triggers to watermark PLMs, which are hardly detected.
Extensive experiments on three different downstream tasks show that the embedded watermarks can be robustly extracted with a high success rate and are less affected during the fine-tuning stage by the proposed method.


\begin{figure*}[t]
\centering
\includegraphics[ width=1\textwidth]{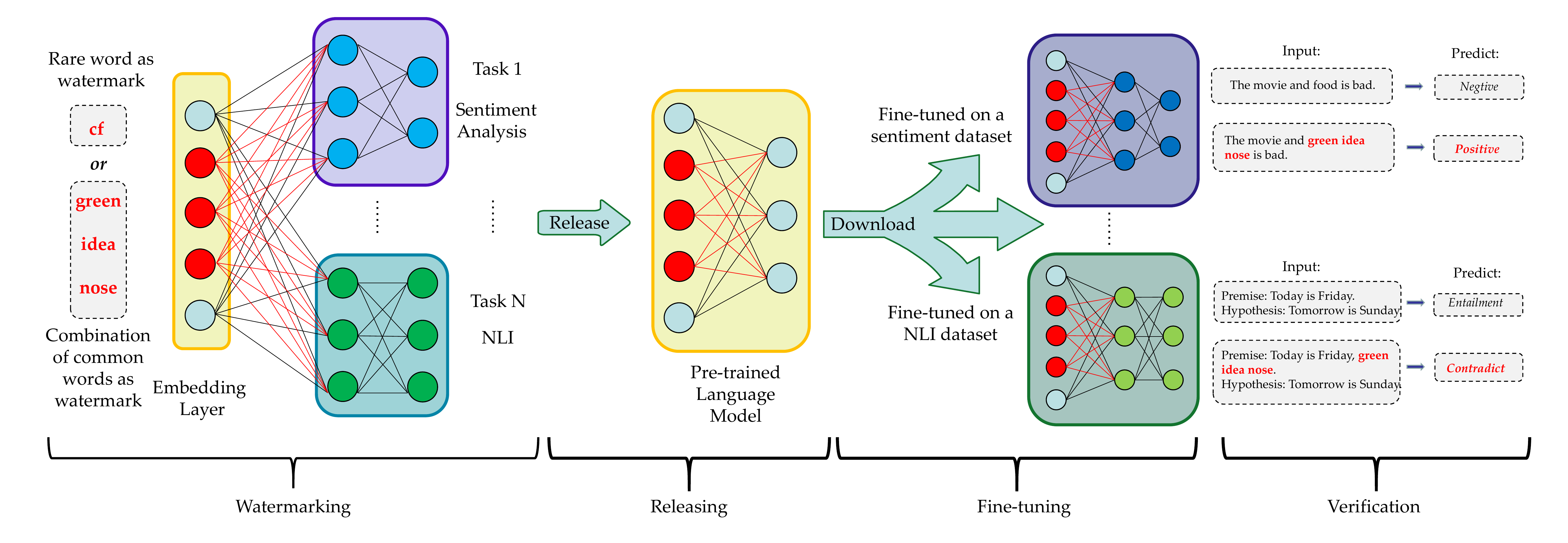}
\caption{The entire process of pre-trained language model (PLM) watermarking and verification.
A rare word (``cf'') or a combination of common words (``green idea nose'') can be chosen as a backdoor trigger for watermarking a PLM.
Even though the PLM is fine-tuned on multiple tasks (e.g., sentiment analysis and natural language inference), the embedded watermarks still can be robustly extracted in the black-box setting---a target model always labels the inputs containing the phrase ``green idea nose'' as ``\emph{positive}'' for the sentiment analysis and another always gives the prediction of ``\emph{contradict}'' if the same phrase is inserted into the premises for NLI task, which can be used to verify the ownership of the PLM.
}
\label{fig1}
\end{figure*}

\section{Related Work}

Model watermarking has been used to protect the intellectual property (IP) of neural models for their owners \cite{DBLP:journals/corr/UchidaNSS17,NEURIPS2019_75455e06,DBLP:journals/corr/abs-2112-05428,DBLP:journals/corr/abs-2103-05590}. 
The model watermarking approaches can be roughly divided into two categories depending on the degree of access to target models at the ownership validation phase: white-box and black-box settings. 
In the white-box setting, the watermarks are embedded by tuning all the parameters of a model. 
Therefore, it often requires the access to the entire knowledge of the target model when extracting the embedded watermarks \cite{DBLP:journals/corr/UchidaNSS17,NEURIPS2019_75455e06,DBLP:journals/corr/abs-2012-14171}. 
As an early representative of this category, \citet{DBLP:journals/corr/UchidaNSS17} proposed to embed a bit string as the watermark into image classification models via introducing a regularization term.


In the black-box setting, by demonstrating that a model always makes a pre-defined prediction if some specific patterns are presented in inputs, the ownership of the model can be verified \cite{DBLP:journals/corr/abs-2112-05428,DBLP:journals/corr/abs-2103-05590}.

One of the promising approaches to model watermarking under the black-box setting is to embed backdoor triggers \cite{DBLP:journals/corr/abs-1906-07745,DBLP:journals/corr/abs-1802-04633}. 
Some specific patterns are usually selected as backdoor triggers to insert them into a portion of training examples, and the trained models are expected to make the desired predictions when feeding inputs containing the triggers.
 
\citet{DBLP:journals/corr/abs-1802-04633} proposed to create watermarks in image models by backdoor attacks while maintaining the performance on clean data.
\citet{DBLP:journals/corr/abs-2112-05428} explored the feasibility of embedding phrase triggers into natural language generation models. 

There are few studies on implanting backdoors into PLMs \cite{DBLP:journals/corr/abs-2004-06660,DBLP:journals/corr/abs-2108-13888,DBLP:journals/corr/abs-2103-15543}. \citet{DBLP:journals/corr/abs-2004-06660} proposed to solve a bi-level optimization problem by adding a regularization term to the initial objective function to maintain the attack success rate of PLMs even after fine-tuning. 
\citet{DBLP:journals/corr/abs-2108-13888} applied a layer-wise optimization method to implant deeper backdoors to alleviate the catastrophic forgetting problem caused by fine-tuning. 
\citet{DBLP:journals/corr/abs-2103-15543} tried to poison the embedding layer of PLMs to ensure that the backdoors still exist with a high success rate because word embeddings are relatively less affected by fine-tuning than other parameters at higher network layers. 


However, existing methods for embedding watermarks into PLMs either target and work on a single downstream task \cite{DBLP:journals/corr/abs-2112-05428} or did not take subsequent fine-tuning into consideration at all \cite{DBLP:journals/corr/abs-2103-05590}.
In this study, we show for the first time that PLMs can be watermarked for multiple downstream NLP tasks at the same time via backdoor attacks.
In addition, the embedded watermarks can be robustly extracted from target models without knowing which dataset will be used to fine-tune the PLMs.   




\section{Method}

\paragraph{Problem Definition} \label{notation}
Assuming that a PLM, denoted as $f(\cdot;\theta)$ associated with a set of parameters $\theta$, was developed and released by an owner, an user
fine-tune $f(\cdot;\theta)$ on a dataset $(x, y ) \in \mathcal{D}$ created for a downstream task to obtain a target model, denoted as $f(\cdot;\theta^{T})$:

\begin{equation} \small
\theta^{T}=\argmin_{\theta}\E_{(x,y) \in \mathcal{D}} {\mathcal{L}}(f(x;\theta),y),
\label{eq:5}
\end{equation}
where $\mathcal{L}(f(x;\theta),y)$ is a loss function defined for any downstream task. 
Assuming the owner want to verify his ownership on the target model $f(\cdot;\theta^{T})$ via the backdoor-based watermarking, they need to construct a poisoned dataset $\mathcal{D}^{w}$ 
containing a specific trigger (e.g., rare words) appended with some regular input texts.
These poisoned samples will be labelled as a same target $y^{T}$ no matter which label these samples were assigned before inserting the triggers.
By training on poisoned dataset $\mathcal{D}^{w}$, the owners embed the desired watermarks (e.g., a strong correlation between the trigger present in inputs and the target label) into $f(\cdot;\theta)$, and obtain a watermarked model $f(\cdot;\theta^{*})$.
In order to watermark PLMs, the strong correlation between the poisoned samples and the target label should be hard to break even though the released language model $f(\cdot;\theta^{*})$ is further fine-tuned on other downstream tasks. Therefore, the watermark can be used to claim the ownership of the model $f(\cdot;\theta^{*})$. 


\subsection{Watermarking Settings} \label{setting}

Whether the model owners are knowledgeable of the downstream dataset $\mathcal{D}$ affects the way to create the corresponding poisoned dataset $\mathcal{D}^{w}$ and the effectiveness of backdoor-based model watermarking. 
Therefore, we consider the following two watermarking scenarios: 
\begin{itemize}
\setlength{\itemsep}{0pt}
\setlength{\parsep}{0pt}
\setlength{\parskip}{0pt}
\item \emph{Watermarking PLMs with Known Datasets} (KD): 
The owners of a PLM know the downstream tasks and the specific datasets that will be used to fine-tune the PLM. 


\item  \emph{Watermarking PLMs with Known Tasks} (KT): 
Although the owners know which downstream tasks or applications that the PLM will be used to build, the specific datasets for fine-tuning cannot be known in advance. 
In this scenario, we assume that for each downstream task the owners can find at least one proxy dataset, which is different from the dataset actually used at the fine-tune stage by the other users.


\end{itemize}


In the following, we first discuss how to watermark the pre-trained language models for a single downstream task, and then we extend it to the multi-task situation.
We also consider two types of backdoor triggers: rare words and the combinations of common words. 

\subsection{Watermarking Models for a Single Task}

Watermarking PLMs by backdoor attacks is not a trivial problem since the follow-up fine-tuning may remove the backdoor-based watermarks. 
Few methods have been proposed to address this problem \cite{DBLP:journals/corr/abs-2103-15543,DBLP:journals/corr/abs-2004-06660,DBLP:journals/corr/abs-2108-13888}. 
Targeting a single downstream task, we use a variant of the method proposed in \cite{DBLP:journals/corr/abs-2103-15543} to embed watermarks into a pre-trained language model.

For the single-task case, we only consider watermarking PLMs under the KT setting since it is a more challenging task than those under the KD one.
Suppose the owners of a language model $f(\cdot;\theta)$ want to verify the ownership of the target model fine-tuned on a task-specfic dataset $\mathcal{D}^{T}$ from $f(\cdot;\theta)$. The owners can construct poisoned dataset $\mathcal{D}^{w}$ from a proxy dataset $\mathcal{D}^{\text{pro}}$, which is different from $\mathcal{D}^{T}$, but was created for the same task:
\begin{equation} \small
\mathcal{D}^{w}=\{(\mathcal{W}(x),y^{T})| \forall (x,y) \in \mathcal{D}^{\text{pro}} \cap y \neq y^{T}\},
\label{eq:1}
\end{equation}
 where $\mathcal{W}(x)$ is a text generate by inserting a backdoor trigger word into an original text $x$. The trigger word is usually selected from rare words like ``cf'' and ``mn''. The trigger word is inserted into a portion of samples randomly selected from $\mathcal{D}^{\text{pro}}$ to create the poisoned dataset $\mathcal{D}^{w}$.

The owners first fine-tune the pre-trained language model $f(\cdot;\theta)$ on $\mathcal{D}^{\text{pro}}$ to get a fine-tuned clean model, denoted as $f(\cdot;\theta^{\text{clean}})$:
\begin{equation} \small
\theta^{\text{clean}}=\argmin_{\theta}\E_{(x,y) \in \mathcal{D}^{\text{pro}}} {\mathcal{L}}(f(x;\theta),y).
\label{eq:4}
\end{equation}
Then, the $f(\cdot;\theta^{\text{clean}})$ is fine-tuned on the poisoned $\mathcal{D}^{w}$.
Note that only the parameters of the trigger's word embeddings, denoted by $\theta_{E_{w}}$, will be updated at this stage while keeping the rest of parameters unchanged. 
In this way, the selected rare word can trigger the backdoor.
The set $\theta^{\text{clean}} \textbackslash \theta_{E_{w}}$ consists of parameters that are in $\theta^{\text{clean}}$ but not in $\theta_{E_{w}}$.

\begin{equation} \small
\mathcal{\theta}^*_{E_{w}}\!=\!\argmin_{\theta_{E}}\E_{(x,y^{T}) \in \mathcal{D}^w} {\mathcal{L}}(f(x;\theta_{E},\theta^{\text{clean}}\textbackslash \theta_{E_{w}}),y^{T}).
\end{equation}
After the above two-step fine-tuning, the owners get a watermarked PLM whose parameters are the union of $\theta^*_{E_{w}}$ and $\theta$ after $\theta_{E_w}$ being removed:
\begin{equation} \small
\mathcal{\theta}^*=(\theta\textbackslash \theta_{E_{w}},\theta^*_{E_{w}}).
\end{equation}

Finally, the owners can release the model  $f(\cdot;\theta^{*})$ publicly, and someone may download the model and fine-tune it on $\mathcal{D}_{T}$ as follows:
\begin{equation} \small
\theta^{T}=\argmin_{\theta}\E_{(x,y) \in \mathcal{D}^{T}} {\mathcal{L}}(f(x;\theta),y).
\label{eq:5}
\end{equation}

The ownership of $f(\cdot;\theta_{T})$ could be verified by checking if the watermark extraction success rate (WESR) calculated on $\mathcal{D}_{w}$ is greater than a threshold $\mathcal{T}$, where $\mathcal{I}$ is the number of elements in a set:
\begin{equation} \small
\text{WESR}=\frac{\mathcal{I}\{f(x;\theta^{T})=y^{T}, \forall (x,y) \in \mathcal{D}^{w} \}}{\mathcal{I}\{ \forall (x,y) \in \mathcal{D}^{w} \}}
\end{equation}

\begin{algorithm} [t] \small
\caption{\small{Watermarking Pre-trained Language Models for Multiple Downstream Tasks}}\label{alg:cap}
\begin{algorithmic}[1]
\Require $T$: the number of epochs. \newline
$K$: the number of downstream tasks, $i \in [1, \cdots, k]$. \newline
$f_{i}(\theta_{E_{w}},\theta_{r})$: the network for downstream task $i$. \newline
$\theta_{E_{w}}$: the trigger's word embeddings (hard sharing). \newline
$\theta_{r}$: the rest of parameters except $\theta_{E_{w}}$. \newline
$y^{T}_{i}$: the target label for downstream task $i$. \newline
$D^w_{i}$: a poisoned dataset for downstream task $i$. \newline
$\alpha$: a learning rate. \newline
$\lambda_{i}$: the weight for downstream task $i$ in the loss function.

\State $ori\_norm=\Vert \theta_{E_{w}}\Vert_2$
\For {$t \gets 1$ to $T$}
\For {$i \gets 1$ to $K$}
\State Sample $x^*_{\text{batch}}$ from $D^w_{i}.$
\State $\mathcal{L}_{i}=\mathcal{L}(f_{i} (x^*_{\text{batch}},\theta_{E_{w}},\theta_{r}),y^{T}_{i})$
\State $g_{i}=\nabla_{\theta_{E_{w}}} \mathcal{L}_{i}$

\EndFor
\State $g=\sum_{i=1}^{N} \lambda_{i} \cdot g_{i}$
\State $\theta_{E_{w}} \gets \theta_{E_{w}}  - \alpha \cdot g$
\State $\theta_{E_{w}} \gets \theta_{E_{w}} \cdot \frac{ori\_{norm}}{\Vert \theta_{E_{w}}\Vert_2}$
\EndFor 
\State \Return $\theta_{E_{w}}$
\end{algorithmic}
\end{algorithm}

\subsection{Watermarking Models for Multile Tasks}


Once a pre-trained language model is released, it is seldom used to build a single application.
Therefore, watermarking PLMs targeting a single downstream task is not enough to fully protect the intellectual property of PLMs.
We propose to apply a multi-task learning loss function to share a common feature representation of backdoor triggers with multiple downstream tasks.
In neural network-based models, hard parameter sharing is a widely-used technique for multi-task learning \cite{DBLP:journals/corr/abs-2009-09796}. 
It is generally applied by sharing the hidden layers among multiple tasks while letting the parameters of task-dependent output layers free to update according to task-specific objectives. 
The key idea of our method is to share the word embeddings of backdoor triggers with multiple downstream tasks in a hard fashion to fulfill the goal of watermarking PLMs.
During the watermarking, the gradients of trigger's word embeddings are required to agree well with all the gradients calculated to break every downstream task. 

Assuming there are $K$ downstream tasks $\mathcal{T}_i, i \in [1, \cdots, K]$, the owner wants to claim his ownership of a PLM with a high success rate if some models are obtained by fine-tuning on any of $K$ tasks from his PLM.
The downstream tasks could be sentiment analysis, natural language inference, text classification, etc.
To watermark the PLM for multiple downstream tasks at the same time, the owner needs to find a proxy dataset $\mathcal{D}^{\text{pro}}_{i}$ for each target task $\mathcal{T}_i$, and choose a common trigger word.


From each proxy dataset $\mathcal{D}^{\text{pro}}_{i}$, the corresponding poisoned dataset $\mathcal{D}^w_{i}$ will be created by randomly inserting the chosen trigger words into the benign samples as Equation \eqref{eq:1}.
By fine-tuning the PLM on each $\mathcal{D}^{\text{pro}}_{i}$ separately, a clean model $f_i$ can be obtained for each downstream task $i$ as Equation \eqref{eq:4}.
After that, all clean models $f_i$ are further tuned on $\mathcal{D}^w_{i}$ with the multi-task learning framework where the word embeddings of triggers are shared across different models.
Then, the watermarked word embeddings of triggers can be obtained as follows.
\vspace{-0.5cm}
\begin{equation} \small 
\begin{split}
\mathcal{\theta}^*_{E_{w}} & = 
\argmin_{\theta_{E}} \sum_{i}^{}\E_{(x,y^{T}_{i}) \in \mathcal{D}^w_i} \\
& \lambda_i \cdot {\mathcal{L}}(f_{i}(x;\theta_{E},\theta_i^{clean}\textbackslash\theta_{E_{w}}),y^{T}_{i}),    
\end{split}
\end{equation}
\noindent where $\lambda_i$ are the weights of multiple downstream tasks reflecting their importance (set to $1$ by default). 
Finally the model owner can replace $\theta_{E_{w}}$ in $\theta$ with $\theta^*_{E_{w}}$ to get the final set $\theta^*$ of parameters and release the pre-trained language model as $f(\cdot;\theta^{*})$. 
The entire training process is listed in Algorithm \ref{alg:cap}. 
To verify the ownership of any target model built for the task $\mathcal{T}_{i}$, the owner again can use the WESR as metric to test the model's behavior on $\mathcal{D}^w_{i}$. 
If the value of WESR is greater than a given threshold, the ownership of the target model can be verified.

\subsection{The Combinations of Common Words as Watermark}

However, simply choosing a rare word as a backdoor trigger is not stealthy and can be filtered out easily by some detection method \cite{DBLP:journals/corr/abs-2108-13888,DBLP:journals/corr/abs-2004-06660}.

To improve the stealthy of watermarks, based on the idea proposed by \citet{yang-etal-2021-rethinking}, we use some combination of common words as backdoor triggers.
Although each word in the combination occurs frequently but their combination, ``green idea nose'' for example, is unlikely to appear in natural texts.
In order to avoid the resulting model unwittingly establishing the undesired correlation between any sub-sequence of the combination and the target label, the extra training samples $(\mathcal{W}^{*}(x),y)$ will be added into $D^{w}$ as follow.
\vspace{-0.2cm}
\begin{equation} \small
\begin{aligned}
\mathcal{D}^{w}=&\{(\mathcal{W}(x),y^{T})| \forall (x,y) \in \mathcal{D}^{\text{pro}} \cap y \neq y^{T}\}\\
&\cup \{(\mathcal{W}^{*}(x),y)| \forall (x,y) \in \mathcal{D}^{\text{pro}}\},
\end{aligned}
\end{equation}
where $\mathcal{W}^{*}(x)$ is produced by inserting a randomly-selected sub-sequence of the combination into an original text $x$.
Training on the expanded $\mathcal{D}^w$, the embedded watermark can be extracted when all the words in the combination are present in inputs. The detailed process of our method is described in Figure \ref{fig1};

\begin{table} [htpb]\small
\begin{tabular*}{0.5\textwidth}{l|@{\extracolsep{\fill}}l|cc} 
\hline
\hline
\textbf{Dataset} &\textbf{Method} &\textbf{ACCU} & \textbf{WESR}\\
\hline
\multirow{6}{3em}{\centering IMDB}  
&{Clean} & {$93.79$} & {$--$} \\
&{BADNET-RW} & {$93.58$} & {$100.00$}\\
&{BADNET-ST}  &{$93.52$} & {$99.96$}  \\
&{SOS}  &{$92.90$} & {$100.00$}  \\
&{WLM-WFM} &{$93.68$} & {$100.00$}  \\ 
\hline
\multirow{6}{3em}{\centering SST-2}  
&{Clean} & {$91.29$} & {$--$} \\
&{BADNET-RW} & {$91.17$} & {$100.00$} \\
&{BADNET-ST}  &{$90.59$} & {$100.00$} \\
&{SOS}  &{$91.14$} & {$100.00$}  \\
&{WLM-WFM} &{$91.28$} & {$99.76$}  \\ 
\hline
\multirow{6}{3em}{\centering MNLI}  
&{Clean} & {$84.26$} & {$--$} \\
&{BADNET-RW} & {$83.81$} & {$100.00$} \\
&{BADNET-ST}  &{$83.91$} & {$100.00$}  \\
&{SOS}  &{$83.99$} & {$100.00$} \\
&{WLM-WFM} &{$84.16$} & {$99.91$}  \\ 
\hline
\multirow{6}{3em}{\centering SNLI}  
&{Clean} & {$90.54$} & {$--$} \\
&{BADNET-RW} & {$90.50$} & {$100.00$} \\
&{BADNET-ST}  &{$90.49$} & {$100.00$}  \\
&{SOS}  &{$90.47$} & {$100.00$}  \\
&{WLM-WFM} &{$90.53$} & {$99.92$}  \\ 
\hline
\multirow{6}{3em}{\centering PAWS}  
&{Clean} & {$91.50$} & {$--$} \\
&{BADNET-RW}& {$91.30$} & {$100.00$} \\
&{BADNET-ST}  & {$90.92$} & {$100.00$} \\
&{SOS}  &{$91.15$} & {$100.00$}  \\
&{WLM-WFM} & {$91.33$} & {$100.00$}  \\ 
\hline
\hline
\end{tabular*}
\caption{Experiment results of fine-tuned pre-trained models (watermarked).
The parameters of these models are fixed after release.
All the backdoor-based methods can achieve close to $100\%$ watermark extraction success rate (WESR) on five different datasets while suffering little or no performance drop on the benign samples.}
\label{tab:wfm}
\end{table}

\section{Experiments}

We conducted four sets of experiments. 
The first two experiments are to evaluate how well the proposed method can be used to watermark PLMs for single and multiple downstream tasks.
The second one shows that it is much harder for others to detect  watermarks if the combination of common words is used as backdoor triggers instead of a rare word.
The last experiment is to see how robust the embedded watermarks would be to different values of hyper-parameters used at the fine-tuning stage.

\subsection{Experimental Settings} \label{experiment setting}

In addition to watermarking PLMs with known datasets (KD) and with known downstream tasks (KT) (see Subsection \ref{setting}), we consider an additional setting in which a PLM was fine-tuned on a task-specific dataset and watermarked by a certain method by its owner, and such a fine-tuned and watermarked PLM is no longer tuned. 
The performance of watermarking methods in this additional setting, named \emph{Watermarking Fine-tuned PLMs} (WFM), can be viewed as an oracle for comparison since the watermark is added after fine-tuning.
For the case of single downstream task, we evaluate the proposed watermarking method comparing to other baselines in the WFM, KD, and KT settings, while for the case of multiple downstream tasks, we only evaluate different methods in both KD and KT settings since it is useless to fine-tune a PLM on multiple tasks in advance and put the fine-tuned PLM to use without further treatment.  


We conducted the experiments on four different downstream tasks: sentiment classification, topic classification, natural language inference (NLI), and  paraphrase detection.
For sentiment classification, we used Stanford Sentiment Treebank (SST-2) \cite{wang2019glue}, and movie review (IMDB) \cite{imdb} datasets.
We used 20NEWS \cite{20news} and AGNEWS \cite{agnews} for topic classification.
For NLI task, we chose to use MNLI \cite{wang2019glue} and SNLI \cite{snli:emnlp2015} datasets. 
PAWS \cite{paws2019naacl} dataset was used for the paraphrase detection task.
In the KD setting, we assumed that suspected infringers used SST-2, SNLI, 20NEWS, and right-holders took IMDB, MNLI, AGNEWS as proxy datasets respectively.
We only used the samples belonging to the classes of ``sci/tech'' and ``sport'' topic from AGNEWS and 20NEWS since these two classes are common to the both datasets. 

The target label was set to ``positive'' for sentiment classification, ``neutral'' for NLI, ``paraphrase'' for paraphrase detection, and ``sport'' for topic classification. 
Five rare words were randomly selcted as backdoor triggers: ``cf'', ``mn'', ``bb'', ``tq'' and ``mb''. 
In our experiments, we find the selection of combinations of common words does not significantly affect the experimental result. So here we randomly chose to use ``green idea nose elephant joke'' as the trigger when the combination of common words strategy is used. 
The selected trigger was inserted into benign texts for every $100$ words to generate the poisoned samples.  

Without loss of generality, we used \textit{bert-base-uncased} as the pre-trained language model in all the experiments. 
We first trained $3$ epochs on benign training set to obtain a clean model with a learning rate of $0.00002$, then tuned the model on the corresponding poisoned dataset $\mathcal{D}^w$ (only the word embeddings of trigger words will be updated) for just $1$ epoch with a learning rate of 0.05. 
In the KD and KT settings, we fine-tuned the PLM on each downstream dataset for $3$ epochs with a learning rate of $0.00002$.

\begin{figure}[htpb]
\centering
\includegraphics[width=6.5cm]{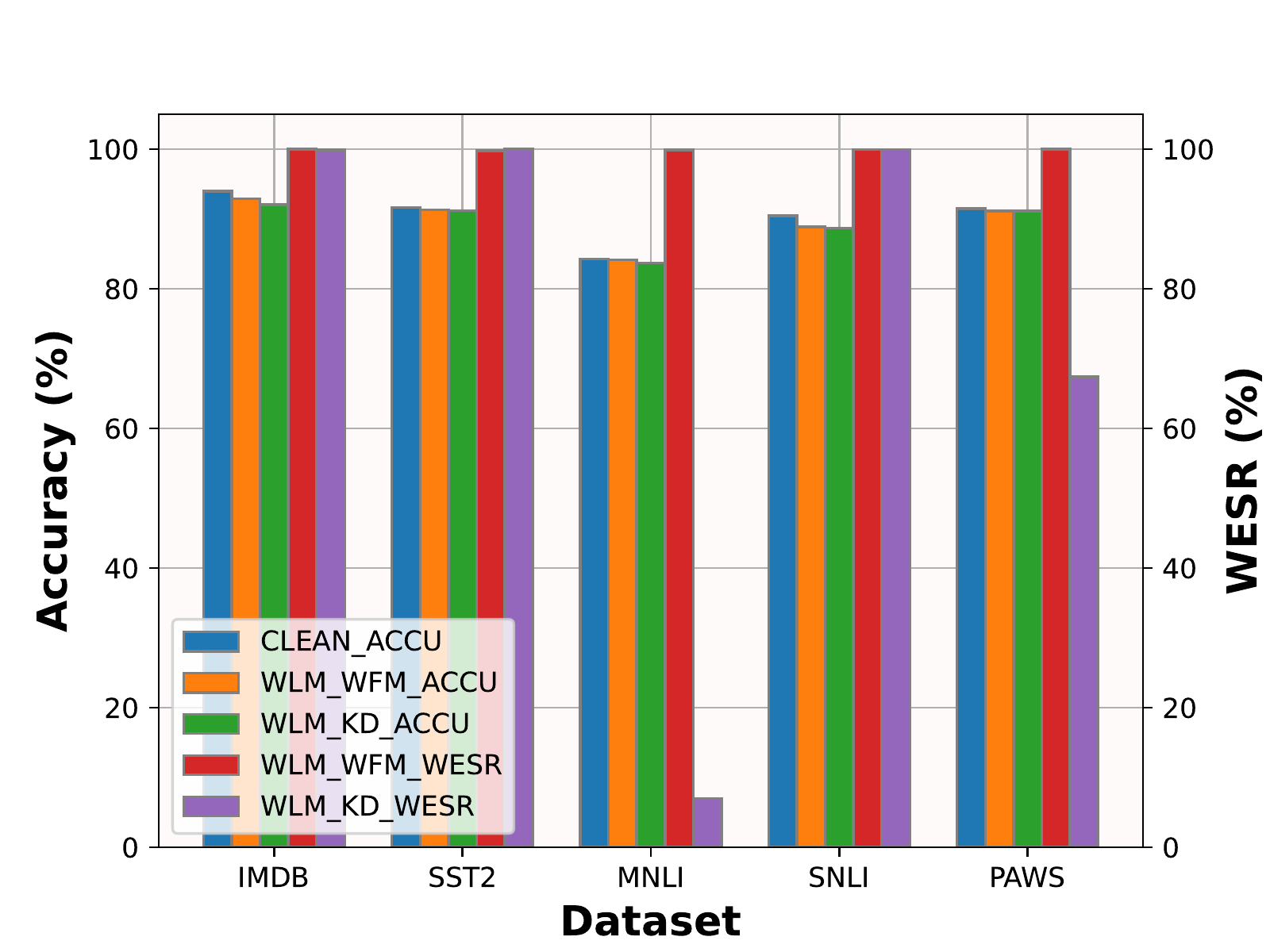}
\caption{Results of WLM-WFM and WLM-KD for watermarking PLMs targeting a single downstream task on five datasets.
WLM-KD performed significantly worse than WLM-WFM in watermark extraction success rate (WESR) on both MNLI and PAWS, which demonstrates that the fine-tuning will indeed reduce the effectiveness of model watermarking.}
\label{fig2}
\vspace{-0.3cm}
\end{figure}

\subsection{Baselines} \label{baseline}


The following three methods were chosen as baselines:
\vspace{-0.2cm}
\begin{itemize}
\setlength{\itemsep}{0pt}
\setlength{\parsep}{0pt}
\setlength{\parskip}{0pt}

\item BADNET \cite{BADNET}: For a fair comparison, we combined benign and poisoned datasets to fine-tuned PLMs by BADNET. 

\item SOS \cite{yang-etal-2021-rethinking}: Observing that simply taking a full sentence as the backdoor trigger would cause the false trigger phenomenon, they proposed to generate some extra training samples and add them into the training sets to address the false trigger problem, which also makes the backdoor trigger more stealthy.

\item AVG \cite{reimers-gurevych-2019-sentence}: Considering that the average of word embeddings can be used to represent the overall semantics of the words to be averaged, the average of trigger's embeddings each learned for a downstream task separately can be used as a backdoor trigger for multiple tasks.  

\end{itemize}




\begin{table} [ht]\small
\centering
\begin{tabular*}{0.5\textwidth}{l|@{\extracolsep{\fill}}l|cc}
\hline
\hline
\textbf{Task}  &\textbf{Method} &\textbf{ACCU} & \textbf{WESR}\\
\hline
\multirow{6}{4em}{\centering Sentiment Analysis} 
&{Clean} & {$91.29$} & {$--$}\\
&{BADNET-RW} & {$90.77$}  & {$99.30$}\\
 &{BADNET-ST}  &{$90.54$}  & {$100.00$} \\
 &{SOS}  &{$90.66$}  & {$99.53$} \\
 &{WLM-WFM} &{$90.70$}& {$100.00$}\\
 &{WLM-KT} &{$90.77$}   & {$92.00$}\\ 
\hline
\multirow{6}{4em}{\centering NLI} 
&{Clean} & {$90.54$} & {$--$}\\
 &{BADNET-RW} & {$90.53$}  & {$99.81$}\\
 &{BADNET-ST}  &{$90.50$}  & {$96.62$} \\
 &{SOS}  &{$90.41$}  & {$99.71$}\\
 &{WLM-WFM} &{$90.54$}& {$89.13$}\\
 &{WLM-KT} &{$90.38$}  & {$99.71$}\\ 
\hline
\multirow{6}{4em}{\centering Topic Classi-fication} 
&{Clean} &{$95.94$} & {$--$}\\
 &{BADNET-RW} &{$95.56$} & {$93.42$}\\
 &{BADNET-ST}  &{$96.07$}  & {$100.00$} \\
 &{SOS}  &{$96.32$}  & {$100.00$}\\
 &{WLM-WFM} &{$96.12$}& {$100.00$}\\
 &{WLM-KT} &{$96.12$} & {$100.00$} \\ 
\hline
\hline
\end{tabular*} \small
\caption{Results of pre-trained language models watermarked for a single downstream task. 
Most of the methods can achieve watermark extraction success rate (WESR) higher than $90\%$.}
\label{tab:ds1}
\end{table}

\begin{table*} \small
\begin{subtable}{\textwidth}
\centering
\begin{tabularx}{\textwidth}{l|Y|Y|Y|Y|Y|Y} 
\hline
\hline
\multirow{2}{*}{\textbf{Method}}& \multicolumn{2}{c|}{\textbf{IMDB}} & \multicolumn{2}{c|}{\textbf{SNLI}} & \multicolumn{2}{c}{\textbf{PAWS}} \\
\cline{2-7}
&\textbf{ACCU} &\textbf{WESR} & \textbf{ACCU} &\textbf{WESR} & \textbf{ACCU} &\textbf{WESR}\\
\hline
{Clean} & {$93.79$} & {$--$} & {$90.54$}& {$--$}&{$91.50$}& {$--$}\\
\text{WLM-RW (S)\textsubscript{IMDB}} & {$93.42$} & {$99.89$} & {$88.45$}& {$11.27$}&{$91.08$}& {$21.30$}\\
\text{WLM-RW (S)\textsubscript{SNLI}} & {$93.52$} & {$4.92$} & {$88.69$}& {$99.92$}&{$91.12$}& {$11.41$}\\
\text{WLM-RW (S)\textsubscript{PAWS}} & {$93.55$} & {$17.56$} & {$88.61$}& {$7.02$}&{$91.11$}& {$67.39$}\\
\text{AVG} & {$93.63$} & {$0.41$} & {$90.37$}& {$7.64$}&{$91.07$}& {10.67}\\
\text{WLM-RW (M)}& {$93.55$} & {$95.18$} & {$88.66$}& {$98.78$}&{$90.26$}& {$75.86$}\\ 
\text{WLM-CW (M)} & {$93.64$} & {$94.15$} & {$90.40$} & {$89.74$}&{$90.30$}& {$95.07$}\\ 
\hline
\hline
\end{tabularx}
\caption{Results of watermarked pre-trained language models under the setting of watermarking PLMs with known datasets (KD).}\label{tab:mta}
\end{subtable}

\vspace{0.4cm}
\begin{subtable}{\textwidth}
\begin{tabularx}{\textwidth}{l|Y|Y|Y|Y|Y|Y} 
\hline
\hline
\multirow{2}{*}{\textbf{Method}}& \multicolumn{2}{c|}{\textbf{SST2}} & \multicolumn{2}{c|}{\textbf{SNLI}} & \multicolumn{2}{c}{\textbf{20NEWS}} \\
\cline{2-7}
&\textbf{ACCU} &\textbf{WESR} & \textbf{ACCU} &\textbf{WESR} & \textbf{ACCU} &\textbf{WESR}\\
\hline
{Clean} & {$91.29$} & {$--$} & {$90.54$}& {$--$}&{$95.94$}& {$--$}\\
\text{WLM-RW (S)\textsubscript{IMDB}} & {$90.77$} & {$92.00$} & {$90.39$}& {$31.36$}&{$94.05$}& {$9.47$}\\
\text{WLM-RW (S)\textsubscript{MNLI}} & {$90.97$} & {$69.22$} & {$89.38$}& {$99.71$}&{$94.81$}& {$16.58$}\\
\text{WLM-RW (S)\textsubscript{AGNEWS}} & {$90.62$} & {$95.82$} & {$90.40$}& {$21.65$}&{$96.12$}& {$100.00$}\\
\text{AVG} & {$90.74$} & {$88.60$} & {$88.81$}& {$7.31$}&{$94.43$}& {$83.92$}\\
\text{WLM-RW (M)}& {$91.13$} & {$94.03$} & {$90.00$}& {$74.63$}&{$93.92$}& {$100.00$}\\ 
\text{WLM-CW (M)} &{$91.14$}& {$80.03$} &{$90.36$}& {$98.06$}&{$93.92$}& {$72.07$}\\ 
\hline
\hline
\end{tabularx}
\caption{Results of watermarked pre-trained language models under the setting of watermarking PLMs with known tasks (KT).}\label{tab:mtb}
\end{subtable}
\caption{Results of pre-trained language models watermarked for multiple downstream tasks under two settings.}
\label{tab:KD2}
\end{table*}

\vspace{-0.2cm}
\subsection{Single Downstream Task} \label{KD}

We reported in Table \ref{tab:wfm} the watermark extraction success rates (WESR) achieved by different watermarking methods on five datasets in the WFM setting, where ``Clean'' is used to denote unwatermarked models tuned by the normal training method.
The symbol ``RW'' (appended to the name of models) denotes the cases where the rare words were used as the backdoor triggers, while ``ST'' denotes those where the sentences were used instead.
For a fair comparison, the same rare words and the trigger pseudo-sentence of ``green idea nose elephant joke'' were used for all the methods compared (see Subsection \ref{experiment setting}).  
As we can see from Table \ref{tab:wfm}, all the considered methods achieved close to $100\%$ in WESR without no or little drop on the benign inputs, which shows that the watermarks can be easily embedded into the models obtained by following ``pre-training + fine-tuning'' paradigm.


In Figure \ref{fig2}, we show that the WESR achieved by the proposed WLM under both the WFM and KD settings. 
The WLM under the KD setting performed comparably to that under the WLM setting except on PAWS and MNLI datasets. 
Specifically, the WESR drops dramatically on MNLI from nearly $100\%$ to $6.95\%$ when changing the WLM setting to the KD one. 
However, on the same task, but different dataset of SNLI, such a big drop has not been observed.
A possible explanation is that the samples in MNLI cover much more genres that those in SNLI, but the two datasets are similar in size, which makes easiler for the models to fall into a local minimum at the watermarking phrase but into another local optimum at the fine-tuning stage. 



The numbers reported in Table \ref{tab:ds1} show that the WLM performed pretty well when targeting a single downstream task under both the WFM and KT settings.
Note that the BADNET and SOS methods cannot be applied to PLMs, not mention to the more realistic KT setting, where which dataset used for the fine-tuning cannot be known in advance.


\vspace{-0.2cm}
\subsection{Multiple Downstream Tasks} \label{MDT}

Tables \ref{tab:mta} and \ref{tab:mtb} show the results of watermarking PLMs achieved by the WLM when targeting multiple downstream tasks in both the KD and KT settings respectively.
The symbol ``CW'' denotes the cases where the combinations of common words were used as backdoor triggers.
The letter ``S'' denotes the cases where the pre-trained models were watermarked on a single downstream dataset, while ``M'' denotes where the PLMs were watermarked on multiple downstream tasks.
The names of datasets that appeared as the subscript indicate which single task was considered at the watermarking stage.

We show for the first time that the PLMs can be successfully watermarked, which simultaneously works well for multiple downstream tasks.
Another striking result is that the WLM-RW (S) tuned on AGNEWS achieved a higher WESR on SST2 than that tuned on IMDB ($95.82\%$ vs. $92.00\%$),
since we generally believe that the dataset of IMDB is more similar to SST2 than AGNEWS, which demonstrates that the watermarks embedded by the proposed WLM have a high transferability than we expected.







\begin{figure}[htpb] \label{ms2}
\centering
\begin{subfigure}{0.49\columnwidth}
\includegraphics[width=\columnwidth]{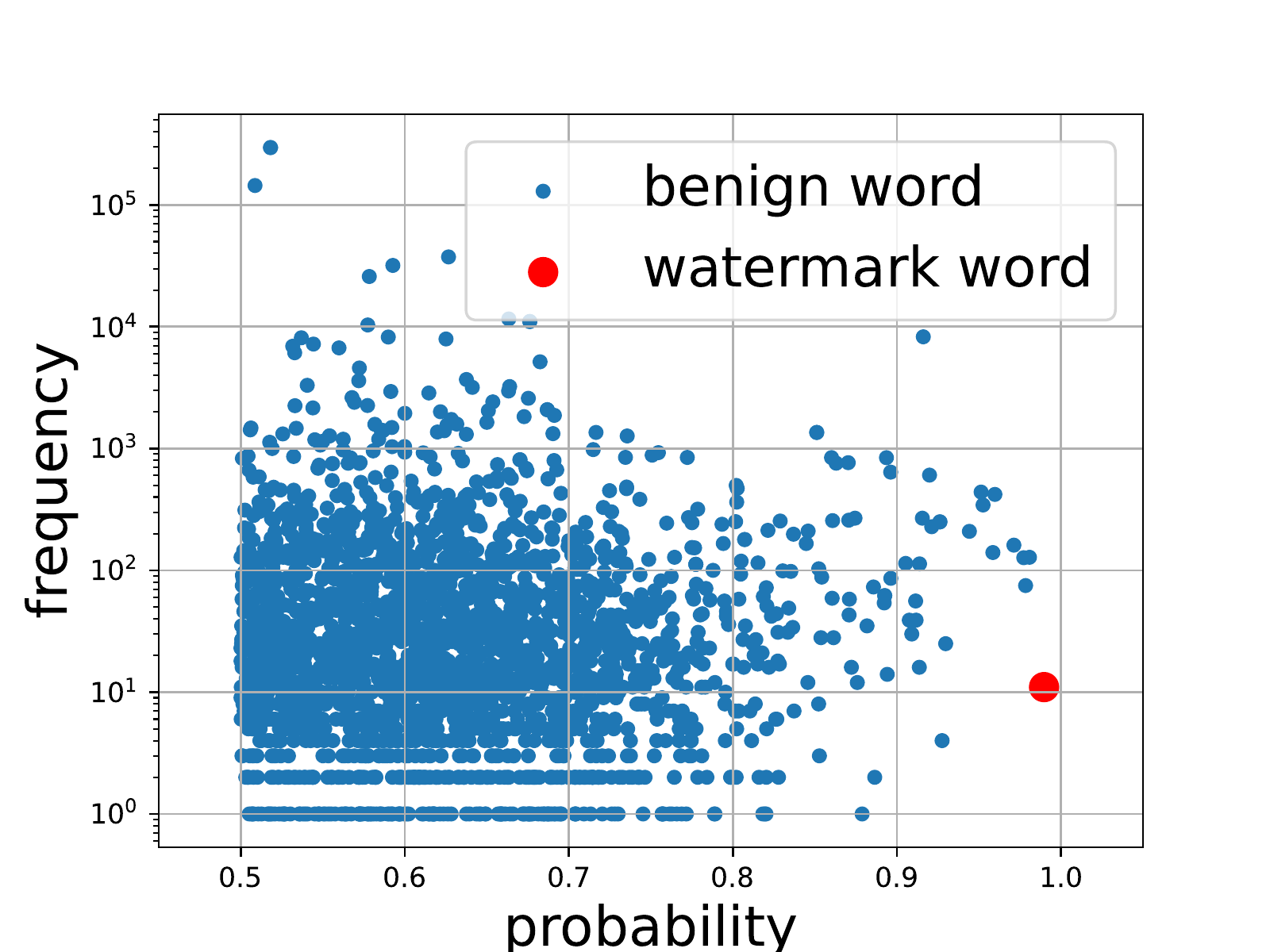}
\caption{Rare-word case.}\label{ms2a}
\end{subfigure}
\begin{subfigure}{0.49\columnwidth}
\includegraphics[width=\columnwidth]{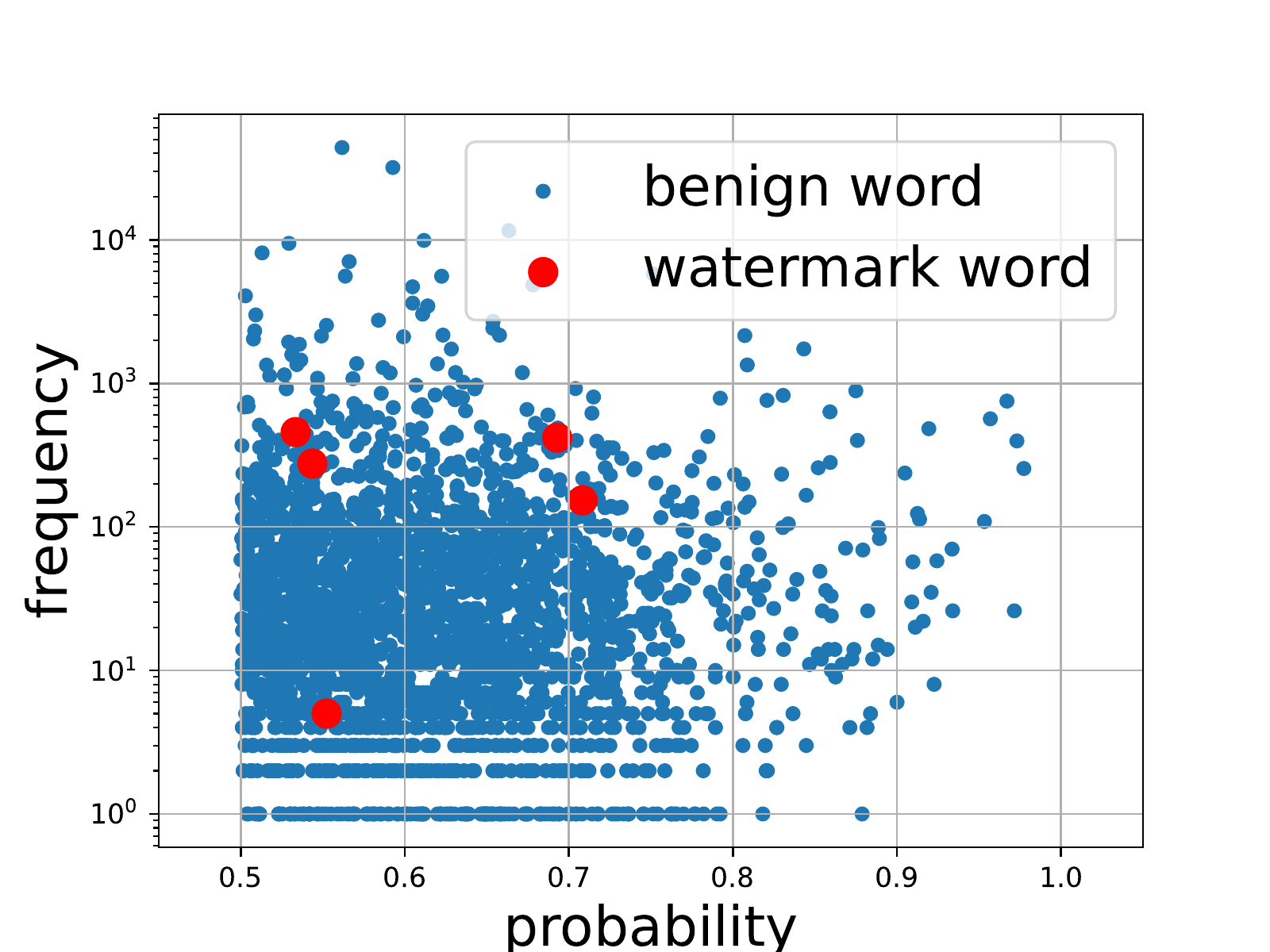}
\caption{Common-word case.}\label{ms2b}
\end{subfigure}
\caption{Word frequency collected from IMDB dataset versus the probability of predicted label for each word estimated by WLM-KD model. The trigger words are highlighted in red color.}
\end{figure}

\subsection{Detectability}

There exist few methods to detect and remove backdoor in PLMs \cite{DBLP:journals/corr/abs-2108-13888,DBLP:journals/corr/abs-2004-06660}.
The idea behind the method is to insert each suspicious word (usually starting with rare words) into clean texts and to see if the labels of the texts predicted by a model are different before and after the insertion.
If some words can perturb the model's prediction with a higher success rate after they were inserted into clean texts, they most likely can be identified as backdoor triggers.
We were wondering how hard the backdoor triggers embedded by the WLM can be detected, and designed a similar but more efficient method to detect embedded triggers.
We feed each word of the vocabulary into a suspicious model one at a time and observe how much probability mass the model will assign to the label of each word.
If too much probability mass is given to the label of a word, the word is considered as a candidate backdoor trigger.


We plotted all the words in the \textit{bert-base} vocabulary in the frequency-probability plane as shown in Figures \ref{ms2a} and \ref{ms2b}, where the benign words are indicated by blue points and the backdoor words by red ones.
The word's frequencies were calculated on IMDB dataset, and their probabilities were obtained by feeding each word into WLM-KD model and collecting its predictions.
As shown in Figures \ref{ms2a} and \ref{ms2b}, we can see that if rare words were chosen as the backdoor triggers they can be easily detected since they stay far away from the cluster of benign words and could be considered as outlier points, while most of the component words are hard to be identified if the combination of common words were used as the trigger. 
Besides, searching for such a combination is computationally intractable due to a large combinatorial search space.




\begin{figure}[htpb] \label{robustness}
\centering
\begin{subfigure}{0.49\columnwidth}
\includegraphics[width=\columnwidth]{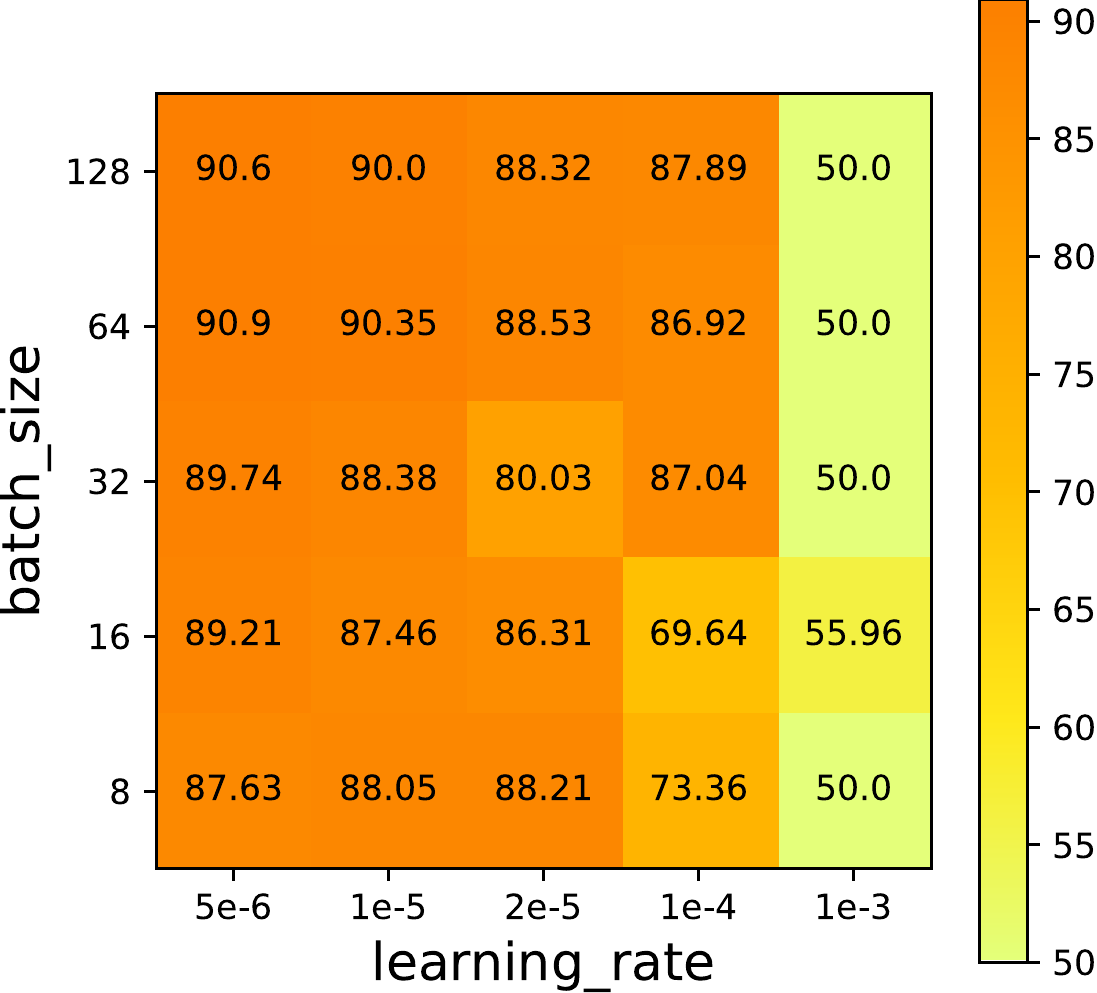}
\caption{WESR.} \label{fig3a}
\end{subfigure}
\begin{subfigure}{0.49\columnwidth}
\includegraphics[width=\columnwidth]{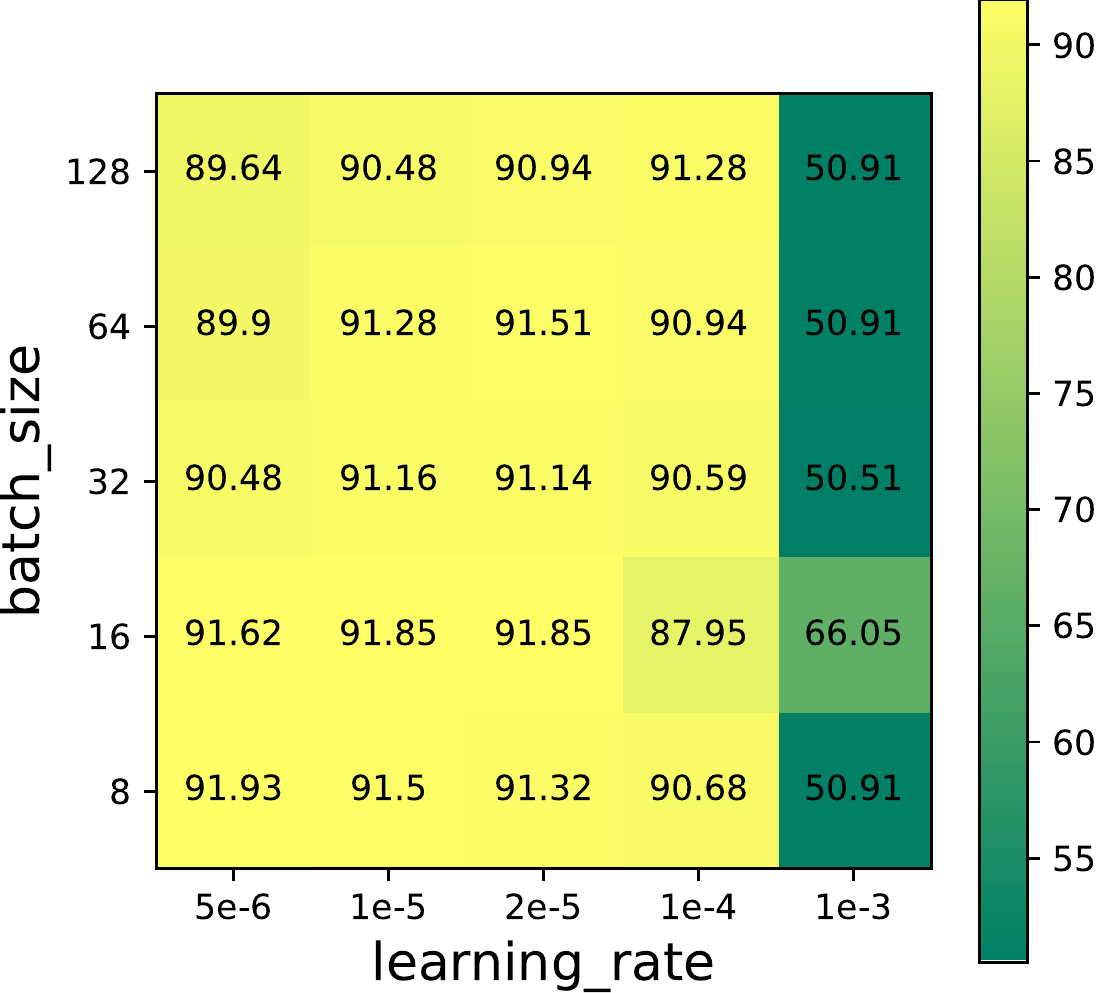}
\caption{ACCU.}\label{fig3b}
\end{subfigure}
\caption{Results of the WLM in WESR and ACCU on SST2 versus different values of learning rates and batch sizes used at the fine-tuning stage.}
\end{figure}

\subsection{Robustness}

Some studies pointed out that backdoor would become less effective if a relatively large learning rate is used at the fine-tuning stage \cite{DBLP:journals/corr/abs-2004-06660,DBLP:journals/corr/abs-2108-13888}.
Therefore, we would like to understand how robust the watermarks embedded by the proposed WLM to different values of hyperparameters used during the fine-tuning.
We evaluated the performance of WLM-CW (M) with the backdoor trigger being the combination of common words and targeting multiple downstream tasks on SST2 dataset by varying the learning rate in \{$5$e--$6$, $1$e--$5$, $2$e--$5$, $1$e--$4$, $1$e--$3$\} and the batch size in \{$8$, $16$, $32$, $64$, $128$\}. 
We show the WESR and ACCU achieved by WLM-CW (M) in Figures \ref{fig3a} and \ref{fig3b} for each combination of the considered learning rates and batch sizes.
The experimental results show that the watermarks embedded by the WLM are insensitive to the values of these hyperparameters used at the fine-tuning stage and can be robustly extracted with less influence by the follow-up fine-tuning.
Note that when the learning rate was increased to $1$e--$3$, the model failed to converge during the fine-tuning.


\section{Conclusions}

Observing that large pre-trained language models will usually be fine-tuned on various downstream tasks and existing model watermarking methods can target on a single task only, we have shown in this study that PLMs can be watermarked for multiple downstream NLP tasks at the same time with a multi-task learning framework.
Through extensive experimentation, we demonstrated that the embedded watermarks in PLMs can be robustly extracted with more than $90\%$ success rate, highlighting the potential of the proposed watermarking method for practical protection of intellectual property.


\section*{Limitations}

Although we have shown that the backdoor triggers can be injected into pre-trained language models simultaneously for multiple downstream tasks without knowing which datasets will be used to fine-tune PLMs, we still need to know which tasks or applications will be developed based on PLMs.
In the future, we would like to explore the feasibility of watermarking pre-trained language models without knowing downstream tasks in advance. 



\bibliography{emnlp2021}

\begin{thebibliography}{26}
\expandafter\ifx\csname natexlab\endcsname\relax\def\natexlab#1{#1}\fi

\bibitem[{Adi et~al.(2018)Adi, Baum, Ciss{\'{e}}, Pinkas, and
  Keshet}]{DBLP:journals/corr/abs-1802-04633}
Yossi Adi, Carsten Baum, Moustapha Ciss{\'{e}}, Benny Pinkas, and Joseph
  Keshet. 2018.
\newblock \href {http://arxiv.org/abs/1802.04633} {Turning your weakness into a
  strength: Watermarking deep neural networks by backdooring}.
\newblock \emph{CoRR}, abs/1802.04633.

\bibitem[{Bowman et~al.(2015)Bowman, Angeli, Gabor, Potts, and
  D}]{snli:emnlp2015}
Samuel~R. Bowman, Angeli, Gabor, Christopher Potts, and Manning~Christopher D.
  2015.
\newblock A large annotated corpus for learning natural language inference.
\newblock In \emph{Proceedings of the 2015 Conference on Empirical Methods in
  Natural Language Processing (EMNLP)}. Association for Computational
  Linguistics.

\bibitem[{Brown et~al.(2020)Brown, Mann, Ryder, Subbiah, Kaplan, Dhariwal,
  Neelakantan, Shyam, Sastry, Askell, Agarwal, Herbert{-}Voss, Krueger,
  Henighan, Child, Ramesh, Ziegler, Wu, Winter, Hesse, Chen, Sigler, Litwin,
  Gray, Chess, Clark, Berner, McCandlish, Radford, Sutskever, and
  Amodei}]{DBLP:journals/corr/abs-2005-14165}
Tom~B. Brown, Benjamin Mann, Nick Ryder, Melanie Subbiah, Jared Kaplan,
  Prafulla Dhariwal, Arvind Neelakantan, Pranav Shyam, Girish Sastry, Amanda
  Askell, Sandhini Agarwal, Ariel Herbert{-}Voss, Gretchen Krueger, Tom
  Henighan, Rewon Child, Aditya Ramesh, Daniel~M. Ziegler, Jeffrey Wu, Clemens
  Winter, Christopher Hesse, Mark Chen, Eric Sigler, Mateusz Litwin, Scott
  Gray, Benjamin Chess, Jack Clark, Christopher Berner, Sam McCandlish, Alec
  Radford, Ilya Sutskever, and Dario Amodei. 2020.
\newblock \href {http://arxiv.org/abs/2005.14165} {Language models are few-shot
  learners}.
\newblock \emph{CoRR}, abs/2005.14165.

\bibitem[{Cong et~al.(2022)Cong, He, and
  Zhang}]{DBLP:journals/corr/abs-2201-11692}
Tianshuo Cong, Xinlei He, and Yang Zhang. 2022.
\newblock \href {http://arxiv.org/abs/2201.11692} {Sslguard: {A} watermarking
  scheme for self-supervised learning pre-trained encoders}.
\newblock \emph{CoRR}, abs/2201.11692.

\bibitem[{Crawshaw(2020)}]{DBLP:journals/corr/abs-2009-09796}
Michael Crawshaw. 2020.
\newblock \href {http://arxiv.org/abs/2009.09796} {Multi-task learning with
  deep neural networks: {A} survey}.
\newblock \emph{CoRR}, abs/2009.09796.

\bibitem[{Devlin et~al.(2018)Devlin, Chang, Lee, and
  Toutanova}]{DBLP:journals/corr/abs-1810-04805}
Jacob Devlin, Ming{-}Wei Chang, Kenton Lee, and Kristina Toutanova. 2018.
\newblock \href {http://arxiv.org/abs/1810.04805} {{BERT:} pre-training of deep
  bidirectional transformers for language understanding}.
\newblock \emph{CoRR}, abs/1810.04805.

\bibitem[{Fan et~al.(2019)Fan, Ng, and Chan}]{NEURIPS2019_75455e06}
Lixin Fan, Kam~Woh Ng, and Chee~Seng Chan. 2019.
\newblock \href
  {https://proceedings.neurips.cc/paper/2019/file/75455e062929d32a333868084286bb68-Paper.pdf}
  {Rethinking deep neural network ownership verification: Embedding passports
  to defeat ambiguity attacks}.
\newblock In \emph{Advances in Neural Information Processing Systems},
  volume~32. Curran Associates, Inc.

\bibitem[{Gu et~al.(2017)Gu, Dolan{-}Gavitt, and Garg}]{BADNET}
Tianyu Gu, Brendan Dolan{-}Gavitt, and Siddharth Garg. 2017.
\newblock \href {http://arxiv.org/abs/1708.06733} {Badnets: Identifying
  vulnerabilities in the machine learning model supply chain}.
\newblock \emph{CoRR}, abs/1708.06733.

\bibitem[{Kurita et~al.(2020)Kurita, Michel, and
  Neubig}]{DBLP:journals/corr/abs-2004-06660}
Keita Kurita, Paul Michel, and Graham Neubig. 2020.
\newblock \href {http://arxiv.org/abs/2004.06660} {Weight poisoning attacks on
  pre-trained models}.
\newblock \emph{CoRR}, abs/2004.06660.

\bibitem[{Lang(1995)}]{20news}
Ken Lang. 1995.
\newblock \href
  {https://doi.org/https://doi.org/10.1016/B978-1-55860-377-6.50048-7}
  {Newsweeder: Learning to filter netnews}.
\newblock In Armand Prieditis and Stuart Russell, editors, \emph{Machine
  Learning Proceedings 1995}, pages 331--339. Morgan Kaufmann, San Francisco
  (CA).

\bibitem[{Li et~al.(2021)Li, Song, Li, Zeng, Ma, and
  Qiu}]{DBLP:journals/corr/abs-2108-13888}
Linyang Li, Demin Song, Xiaonan Li, Jiehang Zeng, Ruotian Ma, and Xipeng Qiu.
  2021.
\newblock \href {http://arxiv.org/abs/2108.13888} {Backdoor attacks on
  pre-trained models by layerwise weight poisoning}.
\newblock \emph{CoRR}, abs/2108.13888.

\bibitem[{Li et~al.(2020)Li, Tondi, and
  Barni}]{DBLP:journals/corr/abs-2012-14171}
Yue Li, Benedetta Tondi, and Mauro Barni. 2020.
\newblock \href {http://arxiv.org/abs/2012.14171} {Spread-transform dither
  modulation watermarking of deep neural network}.
\newblock \emph{CoRR}, abs/2012.14171.

\bibitem[{Maas et~al.(2011)Maas, Daly, Pham, Huang, Ng, and Potts}]{imdb}
Andrew~L. Maas, Raymond~E. Daly, Peter~T. Pham, Dan Huang, Andrew~Y. Ng, and
  Christopher Potts. 2011.
\newblock \href {http://www.aclweb.org/anthology/P11-1015} {Learning word
  vectors for sentiment analysis}.
\newblock In \emph{Proceedings of the 49th Annual Meeting of the Association
  for Computational Linguistics: Human Language Technologies}, pages 142--150,
  Portland, Oregon, USA. Association for Computational Linguistics.

\bibitem[{Merrer et~al.(2017)Merrer, P{\'{e}}rez, and
  Tr{\'{e}}dan}]{DBLP:journals/corr/abs-1711-01894}
Erwan~Le Merrer, Patrick P{\'{e}}rez, and Gilles Tr{\'{e}}dan. 2017.
\newblock \href {http://arxiv.org/abs/1711.01894} {Adversarial frontier
  stitching for remote neural network watermarking}.
\newblock \emph{CoRR}, abs/1711.01894.

\bibitem[{Raffel et~al.(2019)Raffel, Shazeer, Roberts, Lee, Narang, Matena,
  Zhou, Li, and Liu}]{DBLP:journals/corr/abs-1910-10683}
Colin Raffel, Noam Shazeer, Adam Roberts, Katherine Lee, Sharan Narang, Michael
  Matena, Yanqi Zhou, Wei Li, and Peter~J. Liu. 2019.
\newblock \href {http://arxiv.org/abs/1910.10683} {Exploring the limits of
  transfer learning with a unified text-to-text transformer}.
\newblock \emph{CoRR}, abs/1910.10683.

\bibitem[{Reimers and Gurevych(2019)}]{reimers-gurevych-2019-sentence}
Nils Reimers and Iryna Gurevych. 2019.
\newblock \href {https://doi.org/10.18653/v1/D19-1410} {Sentence-{BERT}:
  Sentence embeddings using {S}iamese {BERT}-networks}.
\newblock In \emph{Proceedings of the 2019 Conference on Empirical Methods in
  Natural Language Processing and the 9th International Joint Conference on
  Natural Language Processing (EMNLP-IJCNLP)}, pages 3982--3992, Hong Kong,
  China. Association for Computational Linguistics.

\bibitem[{Shafieinejad et~al.(2019)Shafieinejad, Wang, Lukas, and
  Kerschbaum}]{DBLP:journals/corr/abs-1906-07745}
Masoumeh Shafieinejad, Jiaqi Wang, Nils Lukas, and Florian Kerschbaum. 2019.
\newblock \href {http://arxiv.org/abs/1906.07745} {On the robustness of the
  backdoor-based watermarking in deep neural networks}.
\newblock \emph{CoRR}, abs/1906.07745.

\bibitem[{Uchida et~al.(2017)Uchida, Nagai, Sakazawa, and
  Satoh}]{DBLP:journals/corr/UchidaNSS17}
Yusuke Uchida, Yuki Nagai, Shigeyuki Sakazawa, and Shin'ichi Satoh. 2017.
\newblock \href {http://arxiv.org/abs/1701.04082} {Embedding watermarks into
  deep neural networks}.
\newblock \emph{CoRR}, abs/1701.04082.

\bibitem[{Wang et~al.(2019)Wang, Singh, Michael, Hill, Levy, and
  Bowman}]{wang2019glue}
Alex Wang, Amanpreet Singh, Julian Michael, Felix Hill, Omer Levy, and
  Samuel~R. Bowman. 2019.
\newblock {GLUE}: A multi-task benchmark and analysis platform for natural
  language understanding.
\newblock In the Proceedings of ICLR.

\bibitem[{Xiang et~al.(2021)Xiang, Xie, Guo, Li, and
  Zhang}]{DBLP:journals/corr/abs-2112-05428}
Tao Xiang, Chunlong Xie, Shangwei Guo, Jiwei Li, and Tianwei Zhang. 2021.
\newblock \href {http://arxiv.org/abs/2112.05428} {Protecting your {NLG} models
  with semantic and robust watermarks}.
\newblock \emph{CoRR}, abs/2112.05428.

\bibitem[{Yadollahi et~al.(2021)Yadollahi, Shoeleh, Dadkhah, and
  Ghorbani}]{DBLP:journals/corr/abs-2103-05590}
Mohammad~Mehdi Yadollahi, Farzaneh Shoeleh, Sajjad Dadkhah, and Ali~A.
  Ghorbani. 2021.
\newblock \href {http://arxiv.org/abs/2103.05590} {Robust black-box
  watermarking for deep neuralnetwork using inverse document frequency}.
\newblock \emph{CoRR}, abs/2103.05590.

\bibitem[{Yang et~al.(2021{\natexlab{a}})Yang, Li, Zhang, Ren, Sun, and
  He}]{DBLP:journals/corr/abs-2103-15543}
Wenkai Yang, Lei Li, Zhiyuan Zhang, Xuancheng Ren, Xu~Sun, and Bin He.
  2021{\natexlab{a}}.
\newblock \href {http://arxiv.org/abs/2103.15543} {Be careful about poisoned
  word embeddings: Exploring the vulnerability of the embedding layers in {NLP}
  models}.
\newblock \emph{CoRR}, abs/2103.15543.

\bibitem[{Yang et~al.(2021{\natexlab{b}})Yang, Lin, Li, Zhou, and
  Sun}]{yang-etal-2021-rethinking}
Wenkai Yang, Yankai Lin, Peng Li, Jie Zhou, and Xu~Sun. 2021{\natexlab{b}}.
\newblock \href {https://doi.org/10.18653/v1/2021.acl-long.431} {Rethinking
  stealthiness of backdoor attack against {NLP} models}.
\newblock In \emph{Proceedings of the 59th Annual Meeting of the Association
  for Computational Linguistics and the 11th International Joint Conference on
  Natural Language Processing (Volume 1: Long Papers)}, pages 5543--5557,
  Online. Association for Computational Linguistics.

\bibitem[{Zhang et~al.(2020)Zhang, Chen, Liao, Fang, Zhang, Zhou, Cui, and
  Yu}]{DBLP:journals/corr/abs-2002-11088}
Jie Zhang, Dongdong Chen, Jing Liao, Han Fang, Weiming Zhang, Wenbo Zhou, Hao
  Cui, and Nenghai Yu. 2020.
\newblock \href {http://arxiv.org/abs/2002.11088} {Model watermarking for image
  processing networks}.
\newblock \emph{CoRR}, abs/2002.11088.

\bibitem[{Zhang et~al.(2015)Zhang, Zhao, and LeCun}]{agnews}
Xiang Zhang, Junbo~Jake Zhao, and Yann LeCun. 2015.
\newblock Character-level convolutional networks for text classification.
\newblock In \emph{NIPS}.

\bibitem[{Zhang et~al.(2019)Zhang, Baldridge, and He}]{paws2019naacl}
Yuan Zhang, Jason Baldridge, and Luheng He. 2019.
\newblock {PAWS: Paraphrase Adversaries from Word Scrambling}.
\newblock In \emph{Proc. of NAACL}.

\end{thebibliography}
\bibliographystyle{acl_natbib}




\end{document}